\documentclass[lettersize,journal]{IEEEtran}

\ifCLASSINFOpdf

\else

\fi
\usepackage{soul}
\usepackage{comment}
\usepackage[utf8]{inputenc}
\usepackage{amsmath,amsfonts}
\usepackage{mathtools}
\usepackage{braket}

\usepackage[ruled,vlined,linesnumbered]{algorithm2e}
\let\oldnl\nl
\newcommand{\nonl}{\renewcommand{\nl}{\let\nl\oldnl}}
\usepackage{csquotes}
\usepackage{hhline}
\usepackage{amssymb}
\usepackage{listings}
\usepackage{color}
\definecolor{codegreen}{rgb}{0,0.6,0}
\definecolor{codegray}{rgb}{0.5,0.5,0.5}
\definecolor{codepurple}{rgb}{0.58,0,0.82}
\definecolor{backcolour}{rgb}{0.95,0.95,0.92}

\usepackage{graphicx}
\graphicspath{{./figures/}}
\usepackage{subcaption}
\usepackage{titlesec}

\usepackage{dcolumn}
\usepackage{tabularx}
\usepackage{hyperref}
\hypersetup{
    colorlinks=true,
    linktoc=all,
    linkcolor=black,
    citecolor=magenta,
}
\usepackage{longtable}
\usepackage{braket}
\newcolumntype{C}{>{\centering\arraybackslash}X}
\SetAlFnt{\small}
\usepackage[font=small]{caption}
\begin{document}
\hyphenation{op-tical net-works semi-conduc-tor}

\title{SentiQNF: A Novel Approach to Sentiment Analysis Using Quantum Algorithms and Neuro-Fuzzy Systems}

\author{Kshitij~Dave,
  Nouhaila~Innan,
     Bikash~K.~Behera,
     Zahid~Mumtaz,
    Saif~Al-Kuwari
   and~Ahmed~Farouk

\thanks{K.~Dave is with the Institute of Advanced Research, Gandhinagar, India e-mail: (kshitij.dave@qubrid.com).}
\thanks{N.~Innan N.~Innan is with the quantum Physics and Magnetism Team, LPMC, Faculty of Sciences Ben M'sick, Hassan II University of Casablanca, Morocco. And with the eBRAIN Lab, Division of Engineering, New York University Abu Dhabi (NYUAD), and the Center for Quantum and Topological Systems (CQTS), NYUAD Research Institute, NYUAD, Abu Dhabi, UAE. e-mail: (nouhaila.innan@nyu.edu).}
\thanks{B.~K. Behera is with the Bikash's Quantum (OPC) Pvt. Ltd., Mohanpur, WB, 741246 India, e-mail: (bikas.riki@gmail.com).}
\thanks{Z.~Mumtaz is with the London School of Economics and Political Science, London, UK. e-mail: (Z.mumtaz@lse.ac.uk).}
\thanks{S.~Al-Kuwari is with the Qatar Center for Quantum Computing, College of Science and Engineering, Hamad Bin Khalifa University, Qatar Foundation, Doha, Qatar. e-mail: (smalkuwari@hbku.edu.qa).}
\thanks{A.~Farouk is with the Department of Computer Science, Faculty of Computers and Artificial Intelligence, Hurghada University, Hurghada, Egypt and with the Qatar Center for Quantum Computing, College of Science and Engineering, Hamad Bin Khalifa University, Qatar Foundation, Doha, Qatar. e-mail: (ahmedfarouk@ieee.org).}}

\maketitle

\begin{abstract}
Sentiment analysis is an essential component of natural language processing, used to analyze sentiments, attitudes, and emotional tones in various contexts. It provides valuable insights into public opinion, customer feedback, and user experiences. Researchers have developed various classical machine learning and neuro-fuzzy approaches to address the exponential growth of data and the complexity of language structures in sentiment analysis. However, these approaches often fail to determine the optimal number of clusters, interpret results accurately, handle noise or outliers efficiently, and scale effectively to high-dimensional data. Additionally, they are frequently insensitive to input variations. In this paper, we propose a novel hybrid approach for sentiment analysis called the Quantum Fuzzy Neural Network (QFNN), which leverages quantum properties and incorporates a fuzzy layer to overcome the limitations of classical sentiment analysis algorithms.
In this study, we test the proposed approach on two Twitter datasets: the Coronavirus Tweets Dataset (CVTD) and the General Sentimental Tweets Dataset (GSTD), and compare it with classical and hybrid algorithms. The results demonstrate that QFNN outperforms all classical, quantum, and hybrid algorithms, achieving 100\% and 90\% accuracy in the case of CVTD and GSTD, respectively. Furthermore, QFNN demonstrates its robustness against six different noise models, providing the potential to tackle the computational complexity associated with sentiment analysis on a large scale in a noisy environment. The proposed approach expedites sentiment data processing and precisely analyses different forms of textual data, thereby enhancing sentiment classification and insights associated with sentiment analysis.
\end{abstract}


\IEEEpeerreviewmaketitle
\section{Introduction}\label{QVP:Sec1}
\subsection{An Overview}
\IEEEPARstart{S}{entiment} analysis (SA) is a natural language processing (NLP) technique designed to extract information from textual data, including emotions, attitudes, or opinions expressed by individuals or groups \cite{ganguly2022quantum}.
By analyzing text, SA aims to classify sentiment as positive, negative, or neutral—an inherently challenging task. Various machine learning (ML) algorithms have been successfully applied to perform SA \cite{yang2017survey}. For example, classifiers such as opinion mining, Naive Bayes (NB), J48, BFTree, and OneR, are utilized on Amazon and IMDb movie review data \cite{singh2017optimization,jagdale2019sentiment}. Likewise, SA techniques based on NB, K nearest neighbor (KNN), and random forest (RF) were used for movie reviews \cite{baid2017sentiment}, and algorithms like NB and decision trees (DT) have been employed to assess public opinion \cite{jain2016application}. However, traditional ML algorithms face several challenges in performing accurate and efficient SA tasks. One of the primary difficulties is handling the presence of context-dependent errors, where the interpretation of a word or phrase depends on the surrounding contextual framework \cite{yusof2018review}. Hence, achieving a deep understanding of semantics and pragmatics becomes imperative for accurately determining sentiment. Additionally, sarcasm and irony pose significant challenges, as they often involve conveying meanings opposite to the literal interpretation of the words spoken \cite{maynard2014cares}. Moreover, negation complicates sentiment analysis by using words or phrases that contradict or reverse the meaning of a statement\cite{el2016enhancement}. For example, the sentence ``I do not like this'' expresses a negative sentiment despite containing the positive word ``like''. This complexity is heightened when models require training on multilingual datasets \cite{asif2020sentiment}. Finally, bias in SA can cause algorithms to learn prejudices that result in skewed sentiment classifications \cite{mao2022biases}. This issue becomes even more difficult when dealing with datasets that contain slang words or misspellings, such as those commonly found on social media platforms \cite{neethu2013sentiment}. Quantum computing (QC) harnesses quantum mechanics principles to carry out complex computations more efficiently and offer a speedup over classical computers \cite{18deutsch1992rapid}. In parallel, quantum machine learning (QML) applies the power of quantum algorithms to enhance the capabilities of traditional ML techniques \cite{19biamonte2017quantum}, offering the potential to improve the performance of SA and overcome some of its inherent challenges.

\subsection{Related Work}
Various scholars have explored the use of SA by incorporating classical ML approaches with deep fuzzy networks \cite{13bib_Jasti}. In a particular study, a bidirectional long short-term memory model was combined with a fuzzy system to analyze SA on student feedback gathered from an e-learning platform \cite{14bib_Maryam}. In another study, a comparison was conducted between convolutional neural network (CNN) and fuzzy CNN for SA on various datasets, like Indonesian sentiment Twitter, airline sentiment Twitter, and IMDB film reviews \cite{15bib_Maryam}. Another study introduced a novel semi-supervised learning method, a fuzzy deep belief network with a deep architecture, validated across five SA classification datasets \cite{16bib_Maryam}. Additionally, a modified neuro-fuzzy algorithm was applied to Twitter datasets for SA classification \cite{17bib_Maryam}. 

Beyond classical and fuzzy approaches, recent studies have introduced hybrid quantum-classical and quantum-inspired models to enhance SA performance. A comprehensive review of quantum probability and cognition principles applied to SA is provided in \cite{20liu2023survey}, where quantum-inspired models, such as quantum support vector machines (QSVM) and variational quantum classifiers (VQC), are explored for their potential to outperform traditional methods, despite the challenges posed by real-world datasets. Similarly, the potential of quantum NLP for SA has been demonstrated, showing remarkable accuracy even under noisy quantum environments, thus highlighting the potential of QC for SA tasks \cite{ganguly2022quantum}.  Another innovative method involves complex-valued neural networks, which bridge quantum-inspired models with NLP tasks by effectively handling hidden and complex information for SA was introduced \cite{21lai2023quantum}.  Furthermore, quantum kernels and their potential in classification tasks, particularly for SA, have been extensively studied, with research demonstrating the potential of various quantum techniques such as X and ZX gates for encoding sentiment words in quantum circuits \cite{22sharma2023role}. These techniques were evaluated alongside classical classifiers like support vector machines (SVM) and simultaneous perturbation stochastic approximation, providing valuable insights into effective quantum-based SA methods \cite{23ruskanda2022quantum}. In another study, the quantum-enhanced support vector machine algorithm was used for the representation of sentiment sentences, achieving superior accuracy through circuit parameter optimization and outperforming traditional SVM methods \cite{24ruskanda2023quantum}. The portable quantum language model bridges the gap between quantum and classical machines, demonstrating the potential of quantum pre-training in SA and language modeling \cite{25li2022q}. 
\subsection{Problem Statement}
As clear from the above discussion, the rapid growth and increasing complexity of textual data in SA tasks present considerable challenges for traditional ML approaches, particularly in noisy environments. Existing methods often struggle to maintain both accuracy and efficiency, especially when processing large-scale datasets. While QC offers promising advantages with its unique properties, such as superposition and entanglement, there remains a significant gap in research exploring the integration of QC with advanced SA models under noisy environments and achieving the desired level of accuracy in SA tasks remains a challenge. To address this gap, our study proposes a novel approach that combines quantum algorithms with neuro-fuzzy systems for enhanced SA. Our method's aim is to answer the following key research questions:

\begin{enumerate}
\item Can the proposed Quantum Fuzzy Neural Network (QFNN) model achieve superior sentiment classification performance compared to existing classical, hybrid, and quantum approaches?
\item How does the integration of a fuzzy layer within the QFNN model affect its accuracy, particularly in comparison to hybrid and classical fuzzy (CF) approaches?
\item Does the proposed QFNN model demonstrate improved robustness and efficiency in handling noisy data, which is often common in large-scale sentiment datasets?
\end{enumerate}

By addressing these questions, our study aims to bridge the gap between QC and neuro-fuzzy models, offering a more accurate, efficient, and noise-resilient solution for SA tasks.
\textbf{
\subsection{Novelty of the study}
}
In this study, we present a novel approach to SA tasks by integrating quantum algorithms with neuro-fuzzy systems, leveraging QC advantages like superposition and entanglement. Evaluated on two Twitter datasets (Coronavirus Tweets and General Sentimental Tweets), our model outperformed classical, quantum, and hybrid approaches in sentiment classification accuracy. Additionally, robustness tests against six noise models demonstrated the approach's effectiveness in handling large-scale SA tasks in noisy environments, highlighting its potential for tackling complex textual data on a large scale within noisy environments and providing valuable insights.

\textbf{
\subsection{Contributions of the study}
}

The contributions of this study are as follows:

\begin{itemize}
\item Proposing a novel approach for sentiment detection and classification (SDC) by developing a QFNN, a Hybrid Quantum Neural Network (HQNN), and a Hybrid Fuzzy Neural Network (HFNN). 
\item Introducing a unique method of incorporating fuzziness into a Quantum Neural Network (QNN) by utilizing both Pennylane and Qiskit frameworks, employing predefined embedding and ansatz methodologies.
\item Evaluating the proposed algorithms using key performance metrics, including accuracy, recall, precision, F1 score, false positive (FP) rate, false negative (FN) rate, false discovery (FD) rate, ROC curve, and area under the curve (AUC), and comparing these results with those from existing algorithms in the field.
\item Conducting extensive robustness tests by evaluating the proposed models under six different noise models, demonstrating their resilience and effectiveness in noisy environments, which is critical for real-world applications of sentiment analysis.
\end{itemize}
\begin{figure*}[htbp]
\centering
\begin{subfigure}{\textwidth}
\centering
\includegraphics[width=\linewidth]{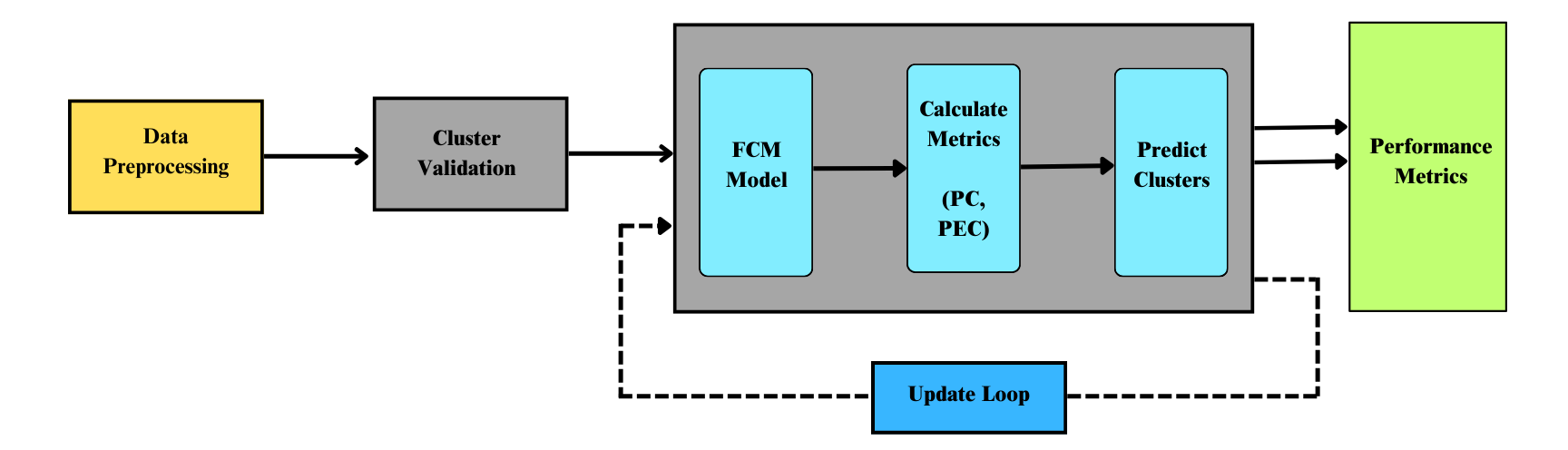}
\caption{}
\label{fig:1a}
\end{subfigure}\hfill
\begin{subfigure}{\textwidth}
\centering
\includegraphics[width=1\linewidth]{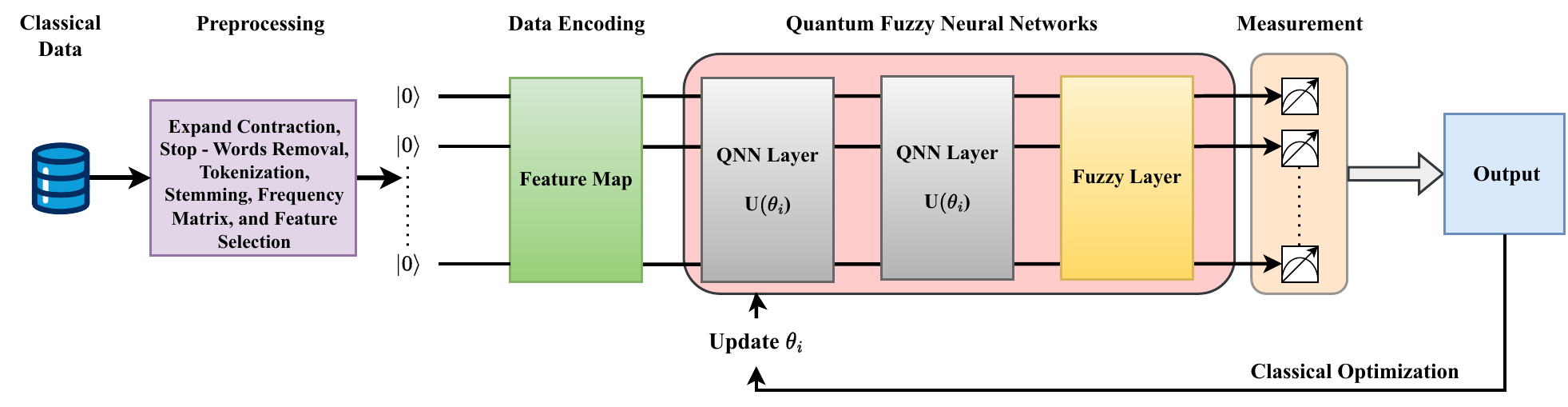}
\caption{}
\label{fig:1b}
\end{subfigure}\hfill 
\caption{The architecture of (a) classical fuzzy (CF) model. (b) QFNN model.}
\label{fig:123}
\end{figure*}

The remainder of the paper is structured as follows: In Section \ref{SecIII}, we introduce the preliminaries, covering the fundamentals of QC, DNN, and fuzzy logic. Section \ref{Sec3} presents the problem formulation along with the methodologies of the proposed algorithms. In Section \ref{SecIV}, we provide the experimental details and results obtained from two specific datasets. Finally, Section \ref{SecV} offers a discussion and conclusion based on the results presented.

\section{Preliminaries}\label{SecIII}
\subsection{Quantum Computing}
Traditional wave-particle classifications prove insufficient within the quantum realm, where subatomic particles find representation through mathematical constructs, particularly the state vector, represented by the generalized wave function $\ket{\Psi}$. These wave functions, existing as vectors within the Hilbert space, can be expressed as,
\begin{eqnarray}
\ket{\Psi} = \sum_{i} c_{i} \ket{\Phi_{i}},
\end{eqnarray} 
where $c_{i}$ represents complex coefficients and $\ket{\Phi_{i}}$ collectively denotes an orthogonal set of basis states.
Entanglement is a quantum phenomenon that occurs when the quantum states of entangled qubits become interconnected in a non-separable way.
Mathematically, for the two-qubit system, an entangled state can be shown by one of the Bell states, $\ket{\Psi} = \alpha\ket{00} + \beta \ket{11}$, where $ \ket{00}$ and $\ket{11}$ are entangled basis states.
In QC, measurements are typically performed in the Z-basis, which is associated with computational basis states $\ket{0}$ and $\ket{1}$. This measurement process can be defined as the linear span of vectors projected onto the Z-basis of the Bloch sphere. It can be expressed as an outer product, denoted as $M_{0} = \ket{0}\bra{0}$ and $ M_{1} = \ket{1}\bra{1}$ corresponding to $\ket{0}$ and $\ket{1}$ respectively. 
In QC, states can be categorized as pure states or as mixed states. For a pure state $\ket{\Psi}$, the corresponding density matrix is defined as $\rho = \ket{\Psi}\bra{\Psi}$. In contrast, for a mixed state composed of multiple pure states $\ket{\Psi_{i}}$, the density matrix is given by,  $\rho = \sum_{i} \ket{\Psi_{i}}\bra{\Psi_{i}}$.
Fundamental single-qubit gates are defined as,
\begin{eqnarray}
X &=&  \ket{0}\bra{1}+\ket{1}\bra{0},Y = -i\ket{0}\bra{1}+i\ket{1}\bra{0},\nonumber\\
Z &=&\ket{0}\bra{0}-\ket{1}\bra{1},I = \ket{0}\bra{0}+\ket{1}\bra{1}, H = \frac{X+Z}{\sqrt2},\nonumber\\ 
S &=&\ket{0}\bra{0}+i\ket{1}\bra{1}, S^{\dagger} = \ket{0}\bra{0}-i\ket{1}\bra{1},\nonumber\\
T &=& \ket{0}\bra{0}+e^{i\frac{\pi}{4}}\ket{1}\bra{1}, T^{\dagger}= \ket{0}\bra{0}+e^{-i\frac{\pi}{4}}\ket{1}\bra{1}.
\end{eqnarray}    
\subsection{Deep Neural Network}
The primary objective of constructing an ML model is to approximate an unknown function that relates input to outputs, with the model's parameters being adjustable to achieve this task effectively. The perceptron, the basic unit of a neural network, adjusts its weights based on input signals to form a linear decision boundary, which can be represented as:

\begin{equation}
f(\sum_{i=1}^{n} (w_{i}x_{i}) + b) = \{1,0\},   
\end{equation}
where $x_{i}$ represents input features, $w_{i}$ signifies the corresponding weights, and $b$ denotes the bias. The function $f$ applies an activation function. The key objective is to adjust the weight values to minimize the discrepancy between the expected output and the model's predictions, thereby reducing the error. 
Popular activation functions include sigmoid, tanh, and ReLU, which are defined as,
\begin{eqnarray}
\sigma(z) &=& \frac{1}{1 + e^{-z}},\\
tanh(z) &=& \frac{e^{z} - e^{-z}}{e^{z} + e^{-z}}, \\
f(z) &=& \max(0,z).
\end{eqnarray}
The training process is essential for evaluating model performance by minimizing the loss function, reducing the error between expected and predicted outputs. Backpropagation computes gradients with respect to the loss function, updating network weights and biases accordingly. The gradients of the loss function L with respect to the network parameters w are computed as:
\begin{equation}
 \frac{dL}{dw} = \frac{dL}{dz}.\frac{dz}{dw},   
\end{equation}
where $\frac{dL}{dw}$, $\frac{dL}{dz}$, and $\frac{dz}{dw}$ are the gradient of the loss for chosen weights, the gradient loss for the output, and the gradient of the output to chosen weights respectively. The update rule for weights and biases is determined by the learning rate ($\alpha$) and the gradient of the loss function for subsequent $W$ parameters ($\nabla$ L(w,b)).
\begin{equation}
    W_{l} = W_{l} - \alpha \frac{\partial L}{\partial W}
\end{equation}

\subsection{Fuzzy Logic}
Fuzzy logic is a mathematical concept that extends traditional binary logic to handle uncertainty and vagueness in a more human-like manner \cite{zadeh1988fuzzy}. 
This makes fuzzy logic more suitable for representing the real world, where information is often imprecise and uncertain.
A fuzzy set $A$ is defined by its membership function $\mu_A(x)$, which assigns a degree of membership $\mu_A(x)$ in the interval $[0, 1]$ to each element $x$ in its universe of discourse.
The membership function can be defined as follows,
\begin{equation}
\mu_A(x) : X \rightarrow [0, 1],
\end{equation}
where $X$ is the universe of discourse for $x$. 
For a set of fuzzy rules ($R_i$), the degree of activation of each rule is calculated as:

\begin{equation}
\alpha_i = \min(\mu_{A_{i1}}(x_1), \mu_{A_{i2}}(x_2), \ldots, \mu_{A_{in}}(x_n)),
\end{equation}
where $\alpha_i$ and $\mu_{A_{ij}}(x_j)$ represent the degree of activation of rule $R_i$ and the membership degree of $x_j$ in the fuzzy set $A_{ij}$ of rule $R_i$ respectively.
The weighted average of the consequent values of all activated rules is calculated to obtain a crisp output $y$ called Defuzzification \cite{KAYACAN201613}:
\begin{equation}
y = \frac{\sum_{i=1}^{N} \alpha_i \cdot \text{Consequent}_i}{\sum{i=1}^{N} \alpha_i},
\end{equation}
where $N$ is the number of activated rules.

\section{Methodology \label{Sec3}}
\subsection{Problem Formulation and Overall Framework}
Consider a dataset with $L$ samples, where the $\gamma$-th sample $X_{\gamma}$ is represented as:
\begin{eqnarray}
    X_{i} = \{(D_\alpha, E_\beta, S_\gamma, S_{t}, T_\phi ), Y_\phi \}.
\end{eqnarray}
Here, $D_\alpha$ corresponds to data cleaning, $E_\beta$ represents expanding, $S_\gamma$ is associated with stemming, $S_{t}$ is related to stopwords, $T_{\phi}$ pertains 
to the target utterance, and $Y_{\phi}$ denotes its corresponding label. 
The term $T_\phi$ can be understood as belonging to the Hilbert space $H^{{l_\phi}.{d_\phi}}$, where $l_{\phi}$ signifies the sequence length of the target utterances and $d_{\phi}$ represents the dimensions of contextual and textual features.
Let $T_\phi$ represents the target utterance, and let the terms $n$ in $T_\phi$ denoted as $t_{i}$, where $i = 1,2,...,n$. The term frequency (TF) is given by:
\begin{eqnarray}
    TF(t_{i}) = \frac{\text{number of occurrences of $t_{i}$ in $T_{\phi}$}}{\text{total number of terms in $T_{\phi}$}}.
\end{eqnarray}
The inverse document frequency (IDF) for a term $t_{i}$ in the document corpus can be expressed as:
\begin{eqnarray}
    IDF(t_i) = log\left(\frac{\text{Total number of documents}}{\text{Number of documents containing $t_{i}$}}\right).
\end{eqnarray}
This formulation enables the representation of the importance of each term in the target utterance $T_{\phi}$ taking into account its frequency in the document and its importance across the entire corpus.\\
The architecture of CF and QFNN used for SA is shown in Fig. \ref{fig:123}. QFNN consists of three main processes, as shown in Fig. \ref{fig:1b}.
\begin{itemize}
    \item The embedding process involves encoding TF-IDF values into the qubits. This approach enables the transformation of the data values into their corresponding angles, facilitating their representation as complex-valued qubit states. As a result, the information is effectively embedded in the qubit states, allowing for efficient processing within the quantum framework.
    \item In the fuzzy layer of neural network architecture, specialized additional layers of neural network parameters function are introduced as fuzzy layers within the neural network to handle complex data patterns.
    \item A fuzzy measurement process converts encoded values into labeled outputs. The adaptive moment estimation (ADAM) optimizer efficiently updates the FNN parameters by adapting the learning rates for each parameter. Additionally, the mean squared error (MSE) loss function is implemented for the classification task, enabling the network to assess the dissimilarity between predicted and actual values and fine-tune its predictive capabilities.
\end{itemize}
\subsection{Theoretical Advantages of Quantum Fuzzy Neural Network Framework}
The concept of superposition in quantum probability uniquely captures the nuances and complexities inherent in human languages. This is because it allows for a nonlinear fusion of the basis and representation of linguistic components, rendering them more dynamically and flexibly expressed. This feature allows simultaneous exploration of multiple linguistic possibilities and provides an overall analysis of language-based data. In the SA, human language's dynamic nature introduces context-dependent interpretations, necessitating an approach that accounts for varying semantic nuances. The adaptability of the QFNN framework proves its ability to handle flowing and context-sensitive natural language and achieve more accurate predictions and classifications even under conditions characterized by linguistic variability. The following three key propositions can highlight the advantages of our proposed QFNN framework.\\

\noindent\emph{Proposition 1:} {Quantum Probability's versatility in capturing linguistic uncertainty.} 

\noindent\emph{Proof.} Consider $z(x) = re^{i\theta}$ denote a quantum complex probability amplitude associated with event $x$. Using the quantum probability definition, the classical probability of event $x$ is
\begin{eqnarray}
p(x) = |z(x)|^{2} = r^{2}, \ r\in R,\ \theta \in (-\pi,\pi).
\end{eqnarray}
Given $p(x)$, the complex probability amplitude can be expressed as
\begin{eqnarray}
z(x) = \sqrt{p(x)}\ (\cos{\theta}+i\sin{\theta}) = re^{i\theta}.
\end{eqnarray}
This yields distinct $r_{1},r_{2} \in R$ and unique $\theta_{1},\theta_{2} \in (-\pi, \pi)$ such that $z_{1}(x) = r_{1}e^{i\theta_{1}}$ and  $z_{2}(x) = r_{2}e^{i\theta_{2}}$ leading to,
\begin{eqnarray}
|z_{1}(x)|^2 = p(x) = |z_{2}(x)|^2, \ r_{1} = r_{2}
\end{eqnarray}

\noindent\emph{Proposition 2:} {Quantum superposition integrates basis states in a non-linear manner.} 

\noindent\emph{Proof.} Let $z_{1}(w_{1})$ and $z_{2}(w_{2})$ denote the complex probability amplitudes of two basis words $w_{1}$ and $w_{2}$, where $z_{1}(w_{1})$,$z_{2}(w_{2}) \in H ^ {l_{t}d_{t}}$. Considering a compound term, $c \propto (w_{1}w_{2})$, the following is obtained,
\begin{eqnarray}
z_{3}(c) = \alpha z_{1}(w_{1}) + \beta z_{2}(w_{2}),\ \alpha, \beta \in H.
\end{eqnarray}
The probability of compound term $c$ is derived as,
\begin{eqnarray}
p(c) = \alpha^{2}p(w_{1})+ \beta^{2}p(w_{2}) +  2\alpha\beta \sqrt{p(w_{1})p(w_2)} \cos{\theta}.
\end{eqnarray}
Now, let us use a contradiction to demonstrate the non-linearity. If the above equation satisfies $kp(x) = p(kx),$ the contradictory conclusion when considering the case $x = \pi/2$ ultimately refutes the linearity assumption. \\

\noindent\emph{Proposition 3:} {Quantum composition system reflects the relations between individual elements and the entire context.}

\noindent\emph{Proof.} Consider two sequential utterances $u_{i}$ and $u_{j}$ as representations of sentiments. The composite system, represented by the state space $H_{u_{i},u_{j}}$, is formed as a tensor product of the individual state spaces $\ket{u_{i}}$ and $\ket{u_{j}}$. If $\ket{w_{1} = (x_{1},x_{2})^{T}}$ and $w_{2} = (y_{1},y_{2})^{T}$, the composite space $H_{u_{i},u_{j}}$ is modulated by the key words and their correlation, reflecting the intricate interplay within the broader context.
\subsection{Complex Valued Utterance Embedding}
The quantum embedding methodology encodes each word value into a set of basis states using a rotational quantum circuit, as shown in Fig. \ref{figcirc}.  
Let \{${\ket{w_{1}},\ket{w_{2}},\ket{w_{3}}....,\ket{w_{n}} }$\} represent the tokenized words within the context of SA. In this framework, each word is encoded as a basis state occupying a space of $2^{n}$ necessary qubits. These states are then mapped into a subspace of the Hilbert space through parameterized gates. Furthermore, the target utterance \{$ X_{i} = {( T_\phi ), Y_\phi}$\}, a key element in SA, can be represented as a complex-valued vector, modeled as a quantum superposition of basis words  \{${\ket{w_{1}},\ket{w_{2}},\ket{w_{3}}....,\ket{w_{n}} }$\}. This representation is expressed as,
\begin{eqnarray}
    \ket{X_{t}} = \sum_{j = 1}^{n}\ z_{j} \ket{w_{j}},\
    z_{j} = r_{j}e^{i\theta_{j}}.
\end{eqnarray}
Where $n$ denotes the number of words in the utterance. The complex probability amplitude $z_{j}$ is written in a polar form, where $r_{j}$ represents magnitude and $\theta_{j}$ indicates phase angle $\theta_{j} \in (-\pi,\pi) $ ensuring ${z_{j}}^ 2 = 1$. 
This novel approach explores language nuances and meaning, ultimately enhancing the accuracy and depth of SA, while offering new perspectives for understanding and classifying textual data.
\subsection{Learning contextual interaction with the fuzzy composition layer}
\begin{figure*}
\hspace{-0.8cm}
    \includegraphics[height=3.5cm,width=19cm]{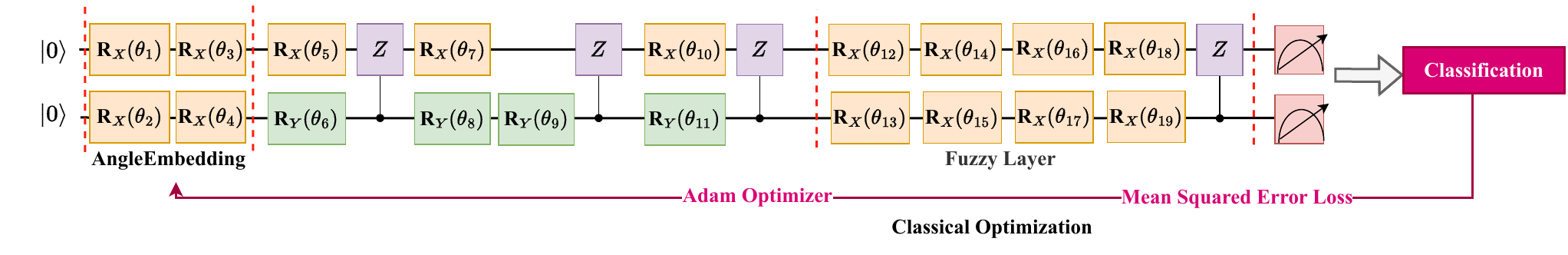}
    \vspace{-0.6cm}
    \caption{The QFNN Circuit.}
    \label{figcirc}
\end{figure*}

The fuzzy composition layer of the proposed QFNN is represented by $R_x$ and $R_y$ gates as shown in Fig.~\ref{figcirc}.
Fuzzy membership functions, represented within the $R_x$ and $R_y$ gates, encode the degree of association between words and their corresponding sentiment polarities. This enables the quantum fuzzy algorithms to capture human language's inherent ambiguity and context-dependent nature. The fuzzy membership function value $\mu(w, S_i)$ for a given word $w$ and sentiment class $S_i$ can be calculated as the weighted sum of individual word associations:
\begin{eqnarray}
\mu(w, S_i) = \sum_{j=1}^{n} F(w_j, S_i) \cdot w_{j},
\end{eqnarray}
where $n$ is the number of words in the text, $w_j$ represents individual words, and $F(w_j, S_i)$ is a function that measures the association between a word $w_j$ and sentiment class $S_i$. The weights $w_{j}$ can be assigned based on the importance or relevance of each word within the context.
Additionally, the overall sentiment score $S_i$ for sentiment class $i$ can be calculated using the fuzzy membership values and fuzzy operators. The combined fuzzy memberships can be expressed as follows,
\begin{equation}
S_i = \mu(w_1, S_i) \otimes \mu(w_2, S_i) \oplus \ldots \oplus \mu(w_n, S_i).
\end{equation}
These operators can be implemented using various mathematical functions like minimum, maximum, or other suitable aggregation functions. The network’s ability to learn contextual interactions increased by introducing $R_x$ and $R_y$ gates into the fuzzy composition, which can be expressed as:
\begin{eqnarray}
R_x(\theta) =
\begin{bmatrix}
\cos{\frac{\theta}{2}} & -i\sin{\frac{\theta}{2}}\\
-i\sin{\frac{\theta}{2}} & \cos{\frac{\theta}{2}}
\end{bmatrix},
R_y(\theta) =\begin{bmatrix}
\cos{\frac{\theta}{2}} & - \sin{\frac{\theta}{2}}\\
\sin{\frac{\theta}{2}} & \cos{\frac{\theta}{2}}
\end{bmatrix}.
\end{eqnarray}
These gates characterized controlled rotations on the Bloch sphere, enabling the network to manipulate qubit states and represent fuzzy membership values with enhanced precision. The controlled-$R_x$ and controlled-$R_y$ gates can be described as,
\begin{eqnarray}
 CR_x(\theta) &=& |0\rangle\langle0| \otimes I + |1\rangle\langle1| \otimes R_x(\theta),\\
 CR_y(\theta) &=& |0\rangle\langle0| \otimes I + |1\rangle\langle1| \otimes R_y(\theta).
\end{eqnarray}
Combining fuzzy logic, $R_x$, and $R_y$ gates in the fuzzy composition layer empowers the quantum fuzzy algorithms to capture contextual nuances and effectively enhance sentiment classification accuracy.

\subsection{Quantum Fuzzy Measurement (QFM)}
The QFM process, a feature map, is crucial in extracting sentiment polarity from the text-embedded quantum state. The QFM process involves projecting a quantum state onto a measurement basis, where the resultant set of distinct vectors shows sentimental classes. The fuzzy layer between measurement and quantum state influences the degree of association between words and their impact on sentiment polarities, eventually providing the flexibility corresponding to nuanced sentiment. The projection operator of the quantum state onto a specific sentiment state $|S_i\rangle$ can be defined as follows,

\begin{eqnarray}
|\psi_i\rangle &= P_i |\psi\rangle, \ P_i = |S_i\rangle\langle S_i|
\end{eqnarray}
With $P_i$ denotes the measurement operator. After combining fuzzy logic the modified projection operator can be defined as,
\begin{eqnarray}
P_i' &=& \mu(w_1, S_i) |S_i\rangle\langle S_i| + \mu(w_2, S_i) |S_i\rangle\langle S_i| \nonumber\\
&&+ \ldots + \mu(w_n, S_i) |S_i\rangle\langle S_i|,
\end{eqnarray}
where $w_1, w_2, ..., w_n$ and $\mu(w_j, S_i)$ represent the words in the text, and the membership value for the jth word about the sentiment class $S_i$ respectively. The probability distribution over sentiment classes from the measurement outcome can be calculated, as $P(S_i) = |\langle\psi|P_i'|\psi\rangle|^2$.
This represents the likelihood of the text belonging to each sentiment class, where the higher probabilities correspond to stronger sentiment associations.

\begin{algorithm}[htbp]
\DontPrintSemicolon
\nonl\textbf{Input:} Training data $X_{\text{train}}$, labels $y_{\text{train}}$, test data $X_{\text{test}}$, learning rate, epochs $N$\;
\nonl\textbf{Output:} Trained quantum fuzz neural network, classification results

Initialize quantum device with 2 qubits\;
Define QFNN with specific layers and gates\;
Create a quantum circuit with 2 qubits\;

\textit{QFNN Layers:}

\textbf{Layer 1:} RX-RY-CZ
\Indp
    Parameters: $\theta_1$ (RX on qubit 0), $\theta_2$ (RY on qubit 1)\;
\Indm

\textbf{Layer 2:} RX-RY-RY-CZ
\Indp
    Parameters: $\theta_3$ (RX on qubit 0), $\theta_4$ (RY on qubit 1), $\theta_2$ (RY on qubit 1)\;
\Indm

\textbf{Fuzzy Layer:}
\Indp
    Parameters: $\theta_5, \theta_6, \theta_7, \theta_8$ (RX and RY gates on both qubits)\;
\Indm

\ForEach{each training epoch \textbf{in} $N$ epochs}
{
Initialize an empty list to store individual batch losses\;
\ForEach {each training batch ($X_{\text{train}}$, $Y_{\text{train}}$)}
    {Initialize random QFNN parameters $\{\theta_1, \theta_2, \ldots, \theta_8\}$\;
    Calculate loss using the QFNN and sigmoid activation\;
    Update QFNN parameters using the ADAM optimizer\;
    Append the batch loss to the list of batch losses \;}}

Evaluate the model on test data ${X}_{\text{test}}$ to obtain accuracy and other metrics\;

Make predictions on test data ${X}_{\text{test}}$ to perform classification\;
\textbf{Return} Trained QFNN, evaluation results\;
\caption{Quantum fuzzy neural network}
\label{algo:qfnn}
\end{algorithm}
\begin{algorithm}[]
\DontPrintSemicolon
\nonl\textbf{Input:} Training data $X_{\text{train}}$ and labels $Y_{\text{train}}$, test data ${X}_{\text{test}}$, learning rate, number of epochs $N$\;
\nonl\textbf{Output:} Trained hybrid quantum neural network model\;
Convert training data and labels to torch tensors\;
Define and create the QNN architecture\;
Initialize feature map with 4 qubits\;
Initialize ansatz with real amplitudes and 4 qubits\;
Create a quantum circuit with 4 qubits\;
Compose the feature map and ansatz into the quantum circuit\;
Create the QNN using the quantum circuit with input gradients enabled\;
Define the torch neural network module\;
Initialize the neural network\;
Define a sequence of layers, including linear and activation layers\;
Connect the QNN using TorchConnector\;
Forward pass through the neural network\;
Concatenate input features and their complements along the last dimension\;
Define batch size and create a data loader\;
Define the ADAM optimizer and MSELoss loss function\;
Set the number of training epochs to 20\;
\ForEach{training epoch of the range of epochs}{
Initialize an empty list to store individual batch losses\;
\ForEach {each training batch ($X_{\text{train}}$, $Y_{\text{train}}$)}
    {
    Initialize gradients to zero\;
    Compute the model's output for input $X_{\text{train}}$\;
    Calculate the loss between the output and the target $Y_{\text{train}}$\;
    Perform backpropagation to compute gradients\;
    Update model weights using the optimizer\;
    Append the batch loss to the list of batch losses\;
    }
}
Evaluate the model on test data ${X}_{\text{test}}$ to obtain accuracy and other metrics\;
Make predictions on test data ${X}_{\text{test}}$ to perform classification\;
\textbf{Return} Trained HQNN, evaluation results\;
\caption{Hybrid quantum neural networks}
\label{algo4}
\end{algorithm}
\begin{algorithm}[]
\DontPrintSemicolon
\nonl\textbf{Input:} Training data $X_{\text{train}}$, labels $y_{\text{train}}$, test data $X_{\text{test}}$, learning rate, epochs $N$\;
\nonl\textbf{Output:} Trained hybrid fuzzy neural network, classification results

Initialize feature map with 4 qubits\;
Initialize ansatz with RealAmplitude and 4 qubits\;
Create a quantum circuit with 4 qubits.\;
Create the QNN using the quantum circuit with input gradient enabled.\;
Define the QNN using TorchConnector.

Insert \textbf{fuzzy layer} of neural network\;

Define the ADAM optimizer and MSELoss loss function\;

Set the number of training epochs to 20\;

\ForEach{training epoch for the range of epochs}{
Initialize an empty list to store individual batch losses\;
\ForEach {each training batch ($X_{\text{train}}$, $Y_{\text{train}}$)}
    {
    Initialize gradients to zero\;
    Compute the model's output for input $X_{\text{train}}$\;
    Calculate the loss between the output and the target $Y_{\text{train}}$\;
    Perform backpropagation to compute gradients\;
    Update model weights using the optimizer\;
    Append the batch loss to the list of batch losses\;
    }
}
Evaluate the model on test data ${X}_{\text{test}}$ to obtain accuracy and other metrics\;

Make predictions on test data ${X}_{\text{test}}$ to perform classification\;

\textbf{Return} Trained hybrid fuzzy neural network, evaluation results\;

\caption{Hybrid fuzzy neural network}
\label{algo:hybrid fuzzy}
\end{algorithm}

\subsection{Quantum Fuzzy Neural Network (QFNN)}
The architecture of the QFNN encompasses several layers, with the extended fuzzy layer being a key component, as shown in Fig.~\ref{fig:1b}. The extended fuzzy layer exhibits characteristics similar to fuzzy logic, a computational paradigm that handles imprecise and uncertain information (Sec. \ref{figcirc}). A thorough analysis is given to understand the layers within the QFNN better.\\
{Input Layer:} The input, representing sentiment text data $X$, undergoes encoding into quantum states using the angle embedding function for quantum processing. This initial encoding is denoted as:
\begin{equation} X \rightarrow \text{angle embedding}(X) = |\phi\rangle, \end{equation}
where $|\phi\rangle$ represents the quantum state encoding the sentiment text.
\begin{eqnarray} 
&&|\phi\rangle \xrightarrow{\text{RX}(\theta_1, \text{qubit 0})} |\psi_1\rangle, \ |\psi_1\rangle \xrightarrow{\text{RY}(\theta_2, \text{qubit 1})} |\psi_2\rangle.  \label{eq30}\\
&&|\psi_2\rangle \xrightarrow{\text{RX}(\theta_3, \text{qubit 0})} |\psi_3\rangle, \ |\psi_3\rangle \xrightarrow{\text{RY}(\theta_4, \text{qubit 1})} |\psi_4\rangle. \label{eq31}\\
&&|\psi_4\rangle \xrightarrow{\text{RX}(\theta_5, \text{qubit 0})} |\psi_5\rangle, \ |\psi_5\rangle \xrightarrow{\text{RY}(\theta_5, \text{qubit 1})} |\psi_6\rangle, \nonumber
\\
&&|\psi_6\rangle \xrightarrow{\text{RX}(\theta_6, \text{qubit 0})} |\psi_7\rangle, \ |\psi_7\rangle \xrightarrow{\text{RY}(\theta_6, \text{qubit 1})} |\psi_8\rangle, \nonumber
\\
&&|\psi_8\rangle \xrightarrow{\text{RX}(\theta_7, \text{qubit 0})} |\psi_9\rangle, \ |\psi_9\rangle \xrightarrow{\text{RY}(\theta_7, \text{qubit 1})} |\psi_{10}\rangle, \nonumber
\\
&&|\psi_{10}\rangle \xrightarrow{\text{RX}(\theta_8, \text{qubit 0})} |\psi_{11}\rangle, \ |\psi_{11}\rangle \xrightarrow{\text{RY}(\theta_8, \text{qubit 1})} |\psi_{12}\rangle.\nonumber\\
\label{eq32}\end{eqnarray}
In the first layer (Eq. \ref{eq30}), quantum variability is introduced by applying rotational-X $(RX)$ and rotational-Y $(RY)$ gates on qubits 0 and 1. These gates, parameterized by $\theta_1$ and $\theta_2$, respectively, produce a degree of uncertainty in the quantum state. In the second layer (Eq. \ref{eq31}), building upon the first layer, it continues to incorporate RX and RY gates, now parameterized by $\theta_3$ and $\theta_4$. In the extended fuzzy layer (Eq. \ref{eq32}), at the core of this architecture, the extended fuzzy layer plays a pivotal role in capturing nuanced sentiment features. This layer comprises multiple repetitions of RX and RY gates on both qubits. These gate sequences introduce a controlled level of imprecision and uncertainty into the quantum state, reflecting the fundamental principles of fuzzy logic. Furthermore, applying controlled-Z $(CZ)$ gates between qubits 0 and 1 following each block creates quantum entanglement, akin to the complex linguistic relationships accommodated by fuzzy logic.
The QFNN is trained by minimizing a cost function that quantifies the difference between predicted sentiment labels and ground truth. This iterative process fine-tunes the parameters ${\theta_1, \theta_2, \ldots, \theta_8}$ to optimize sentiment classification accuracy. The extended fuzzy layer's significance is its unique ability to integrate quantum feature extraction with the ethos of fuzzy logic. The imprecision and capturing of complex sentiment nuances, particularly those shaped by contextual dependencies and ambiguous linguistic expressions, can be managed efficiently by the QFNN (see Algorithm \ref{algo:qfnn}).

\subsection{Hybrid quantum neural network (HQNN)}
A trained HQNN model integrating classical ML with principles of QC is proposed. The training data ${X}_{train}$ and labels ${Y}_{train}$ are initially converted into torch tensors to enable compatibility with the PyTorch deep learning framework. The HQNN architecture incorporates a quantum feature map and a parametrized quantum circuit to facilitate quantum-inspired processing. The feature map encodes input data into quantum states, while the parametrized quantum circuit introduces variational parameters to enable subsequent optimization. These components are combined within a quantum circuit to form the QNN. A TorchConnector is utilized to bridge the classical and quantum computations. For the training process, mini-batch stochastic gradient descent is employed. Each epoch consists of iterations over mini-batches, as detailed in Algorithm \ref{algo4}. In each iteration, the input features are processed by the HQNN algorithm, producing predictions. 
The MSE loss function quantifies the discrepancy between predictions and ground truth labels. The backpropagation algorithm calculates gradients, enabling weight updates through the ADAM optimizer. Consequently, the feature representation is enhanced since the input features with complements are concatenated by $\text{Concatenate}(x, 1-x, \text{dim}=-1)$.
\subsection{Hybrid fuzzy neural network (HFNN)} 
HFNN combines a fuzzy layer within the neural network architecture, which introduces a degree of ``fuzziness'' to the network’s decision-making process and effectively allows it to handle uncertainties and complex patterns within the complex data. It initializes a quantum feature and ansatz, forming the core quantum circuit. A QNN is constructed, where input gradients are enabled to facilitate training. Using the hybridization process, TorchConnector interfaces the QNN with Pytorch. The training process involves using the ADAM optimizer and MSE loss function. Throughout 20 training epochs, the network undergoes iterative weight updates using backpropagation and gradient descent. Upon training completion, the HFNN is evaluated on test data to assess its performance against accuracy and other relevant metrics. The detailed process of HFNN is given by Algorithm \ref{algo:hybrid fuzzy}.

\section{Experimental Results and Discussion\label{SecIV}}

\begin{figure*}
    \centering
    \begin{subfigure}{0.5\textwidth}
        \includegraphics[width=\linewidth]
        {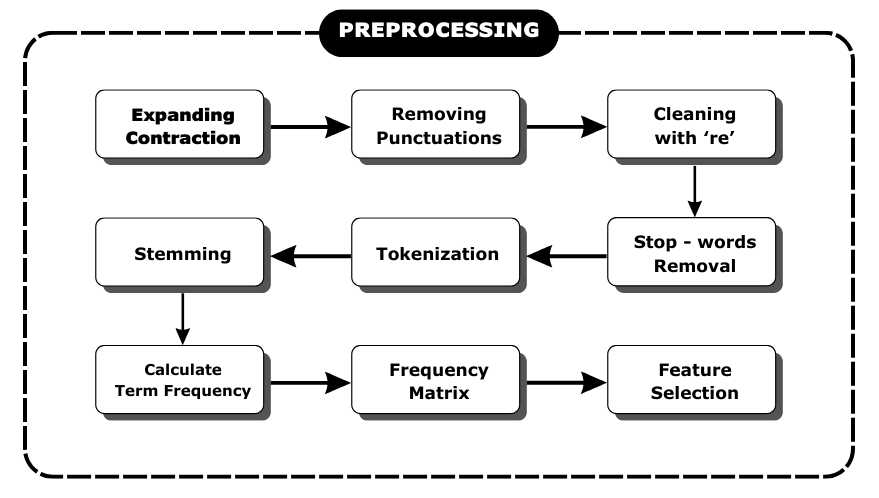}
        \caption{}
        \label{figpreprocess}
        
    \end{subfigure}\hfill
    \begin{subfigure}{0.5\textwidth}
        \includegraphics[width=\linewidth]
        {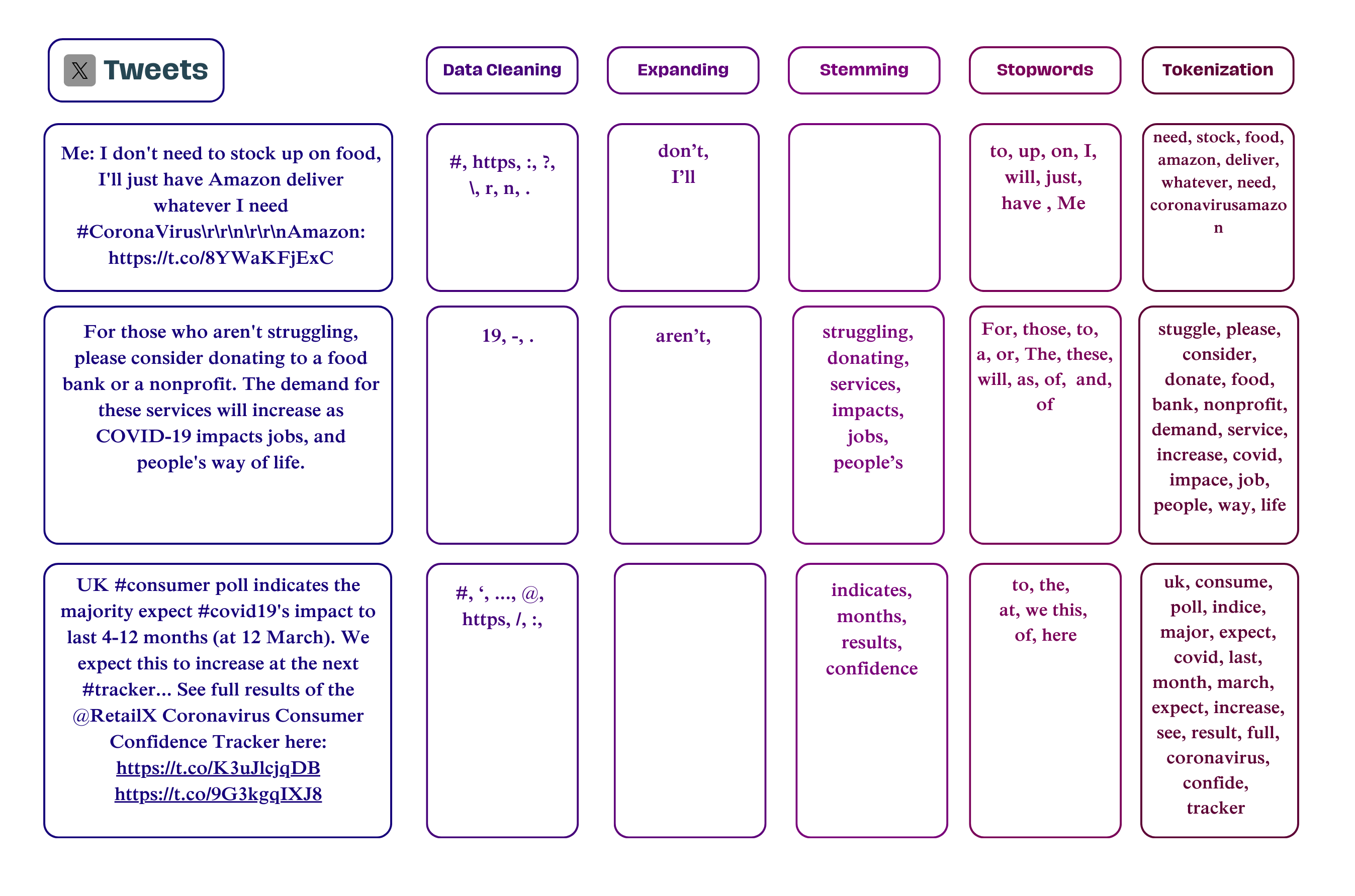}
        \caption{}
        \label{figtweets}
        
    \end{subfigure}\hfill
    \caption{(a) Flowchart illustrating the preprocessing steps for sentiment analysis, and (b) represents an example of preprocessing steps applied to tweets for sentiment analysis.}
\end{figure*}

\begin{figure}[]
\centering
\begin{subfigure}{0.5\linewidth}
\centering
\includegraphics[width=\linewidth]{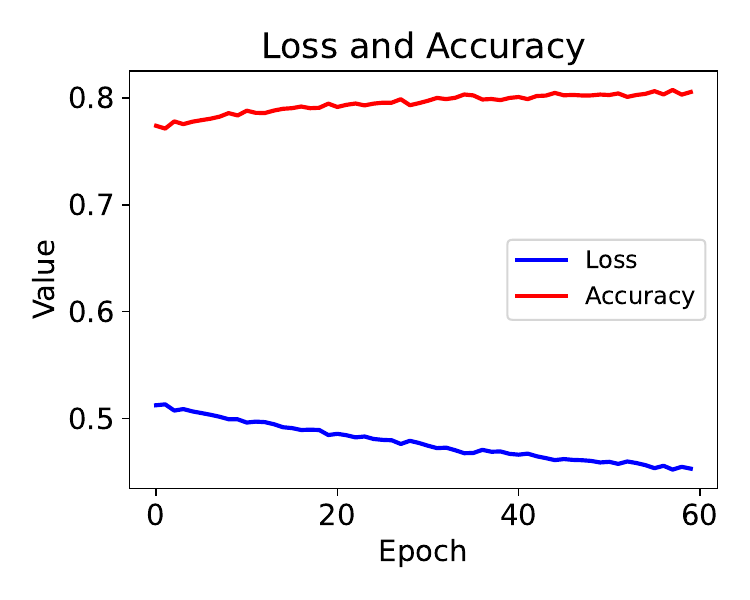}
\caption{}
\label{Fig4a}
\end{subfigure}\hfill
\begin{subfigure}{0.5\linewidth}
\centering
\includegraphics[width=\linewidth]{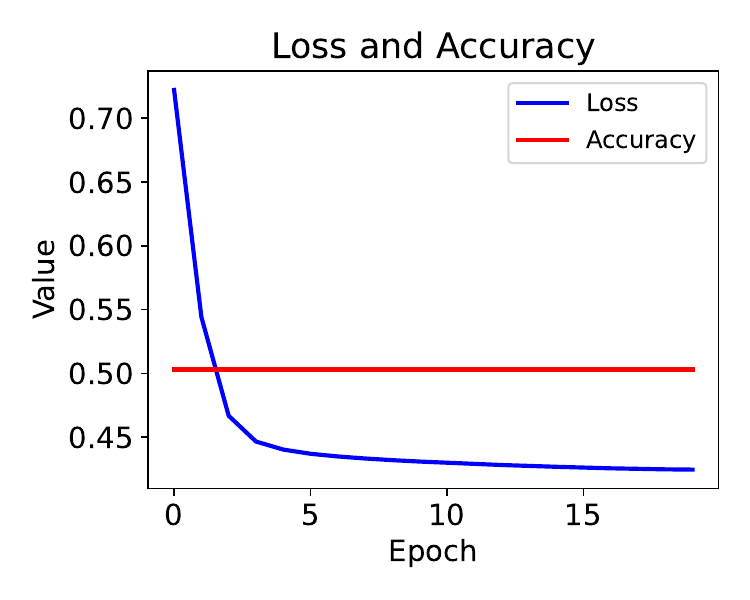}
\caption{}
\label{Fig4b}
\end{subfigure}\hfill
\begin{subfigure}{0.5\linewidth}
\centering
\includegraphics[width=\linewidth]{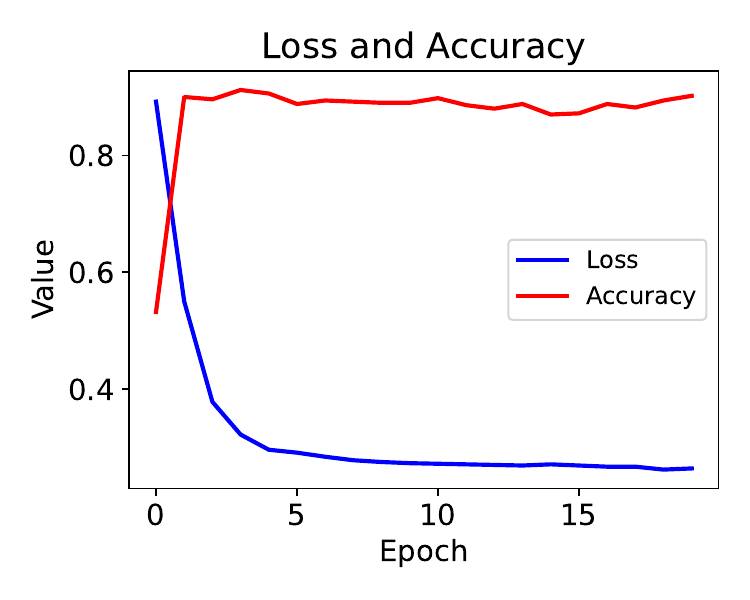}
\caption{}
\label{Fig4c}
\end{subfigure}\hfill
\begin{subfigure}{0.5\linewidth}
\centering
\includegraphics[width=\linewidth]{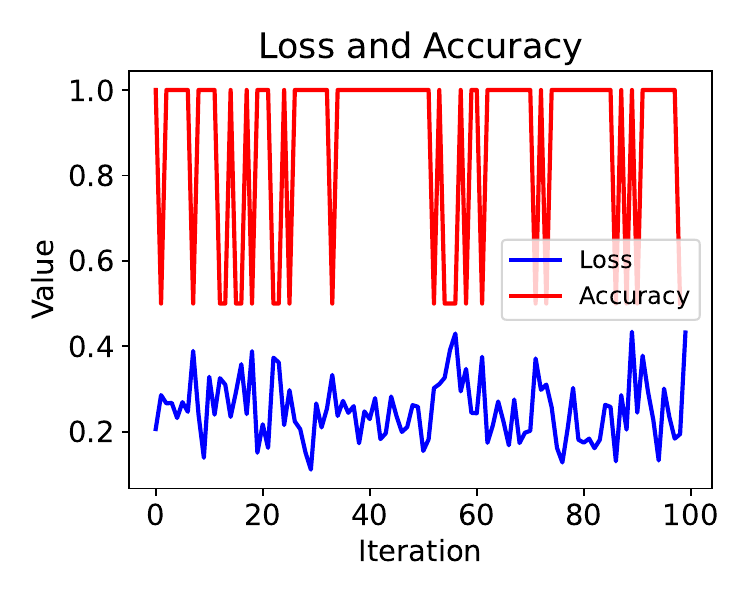}
\caption{}
\label{Fig4d}
\end{subfigure}
\begin{subfigure}{0.5\linewidth}
\centering
\includegraphics[width=\linewidth]{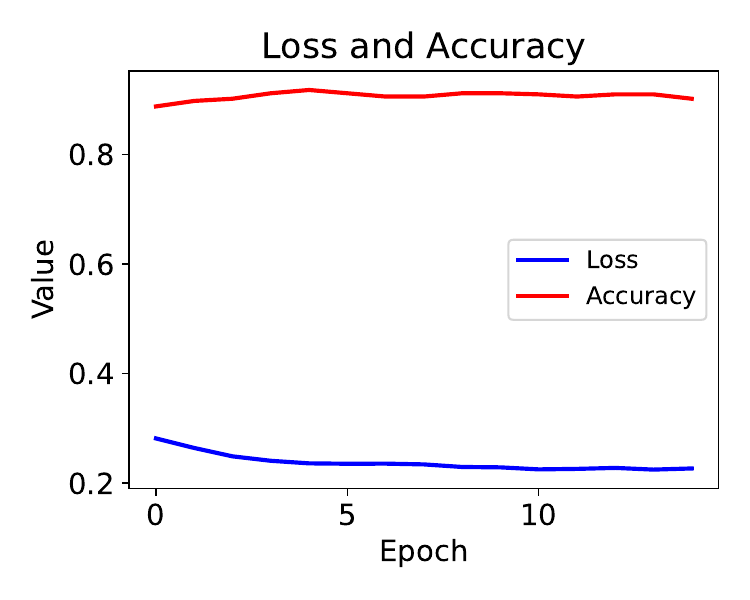}
\caption{}
\label{Fig4e}
\end{subfigure}
\caption{The loss and accuracy of CVTD for (a) ANN, (b) QNN, (c) HQNN, (d) QFNN, and (e) HFNN.}
\label{Fig4-}
\end{figure}

\begin{figure}[htbp]
\centering
\begin{subfigure}{0.5\linewidth}
\centering
\includegraphics[width=\linewidth]{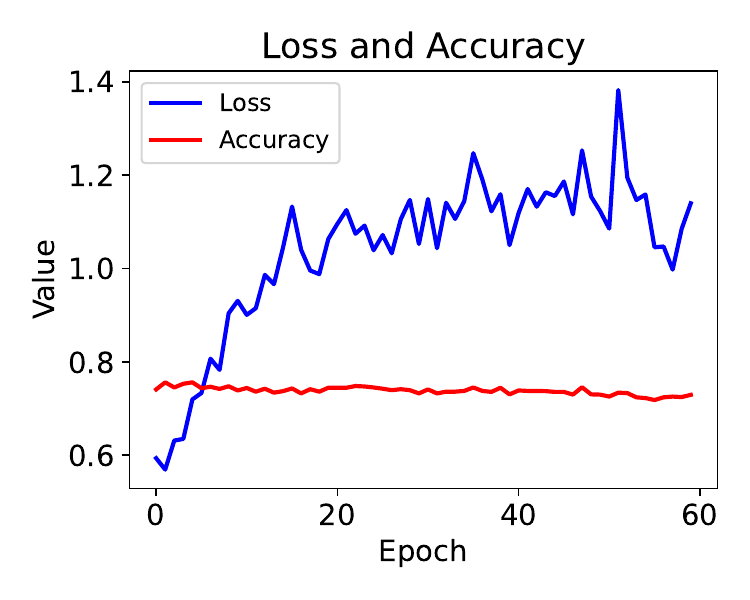}
\caption{}
\label{Fig4a2}
\end{subfigure}\hfill
\begin{subfigure}{0.5\linewidth}
\centering
\includegraphics[width=\linewidth]{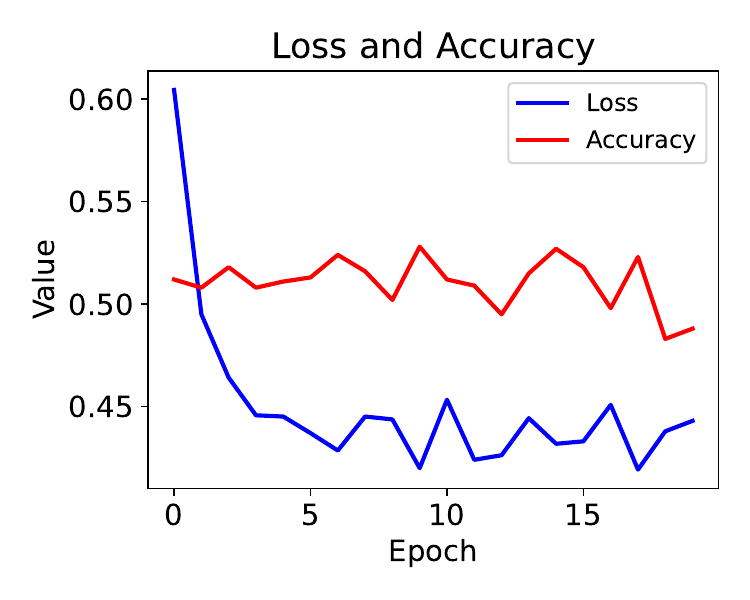}
\caption{}
\label{Fig4b2}
\end{subfigure}\hfill
\begin{subfigure}{0.5\linewidth}
\centering
\includegraphics[width=\linewidth]{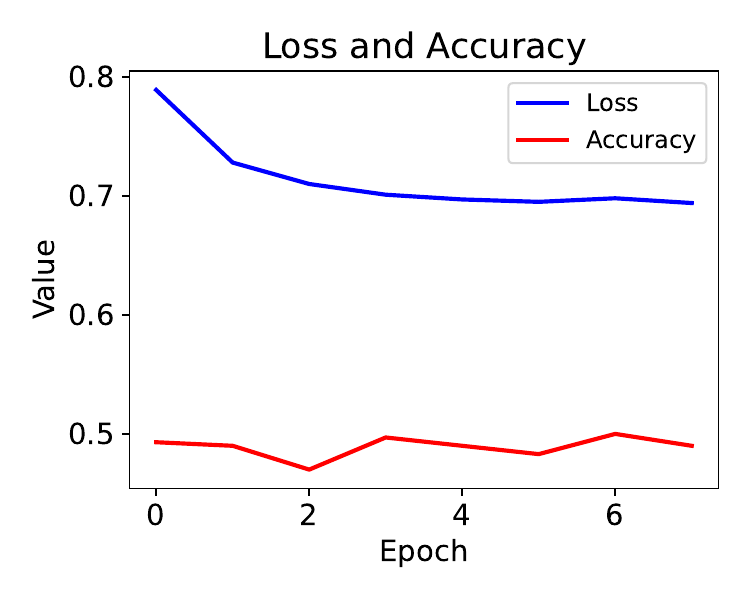}
\caption{}
\label{Fig4c2}
\end{subfigure}\hfill
\begin{subfigure}{0.5\linewidth}
\centering
\includegraphics[width=\linewidth]{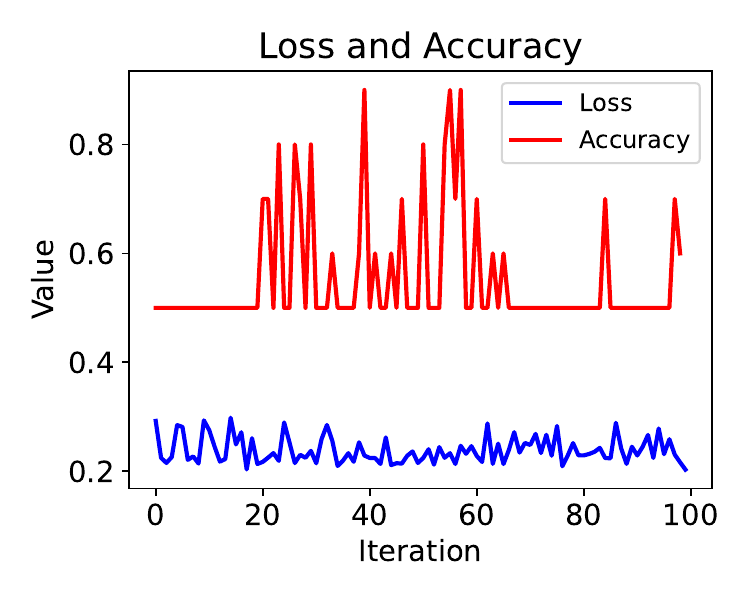}
\caption{}
\label{Fig4d2}
\end{subfigure}
\begin{subfigure}{0.5\linewidth}
\centering
\includegraphics[width=\linewidth]{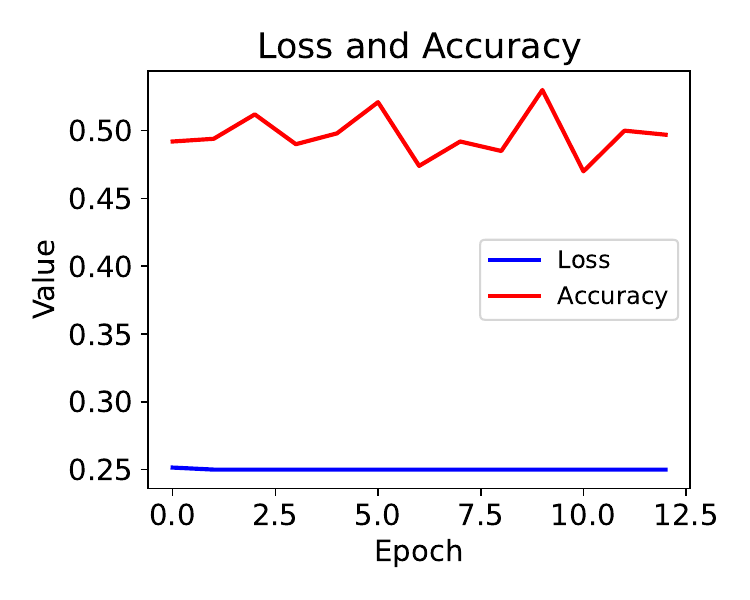}
\caption{}
\label{Fig4e2}
\end{subfigure}
\caption{The loss and accuracy of GSTD for (a) ANN, (b) QNN, (c) HQNN, (d) QFNN, and (e) HFNN.}
\label{Fig4}
\end{figure}

\begin{figure}[!ht]
\centering
\begin{subfigure}{\linewidth}
\centering
\includegraphics[width=0.8\linewidth]
{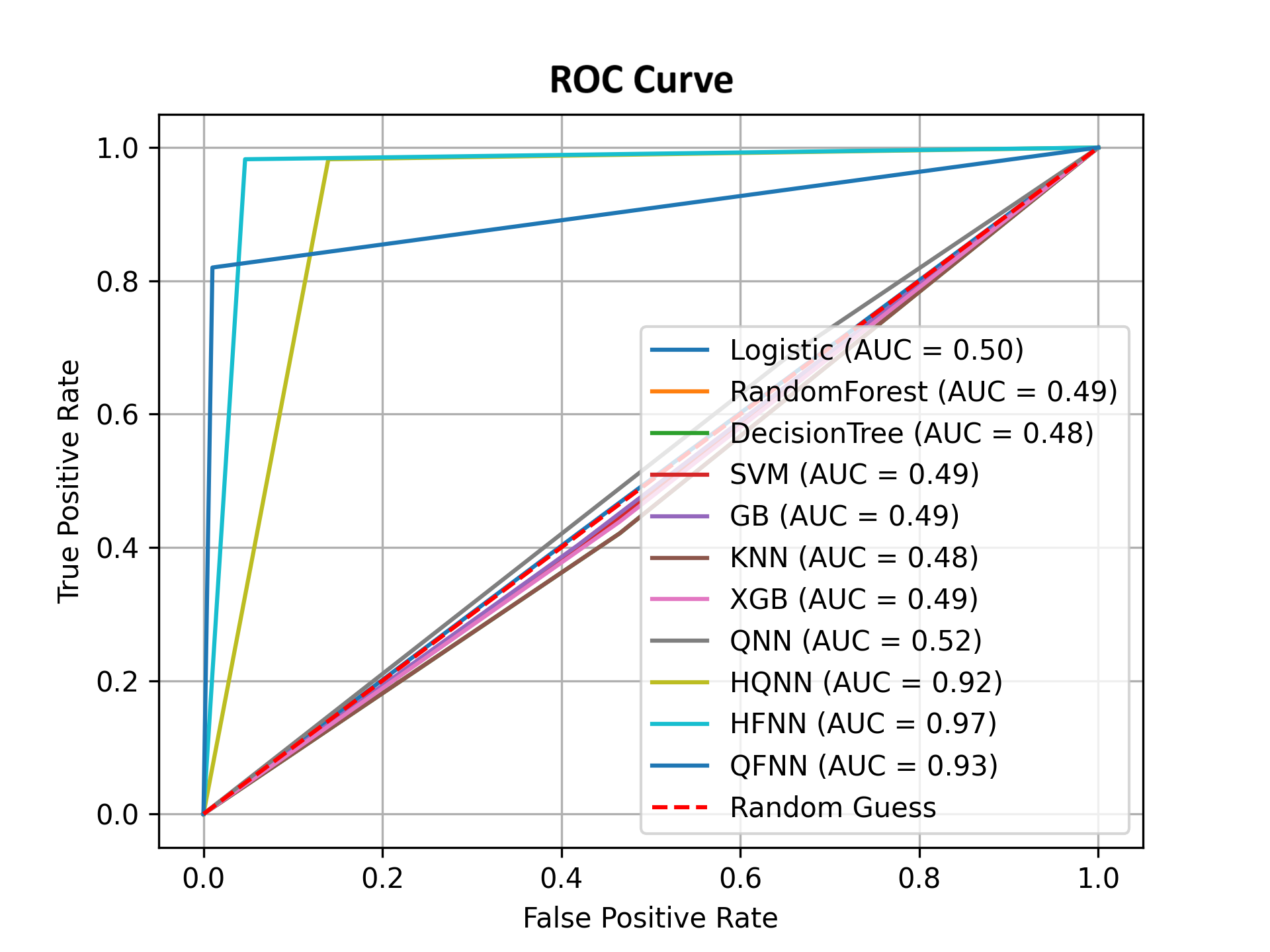}
\caption{}
\label{roc1}
\end{subfigure}\hfill
\begin{subfigure}{\linewidth}
\centering
\includegraphics[width=0.8\linewidth]%
{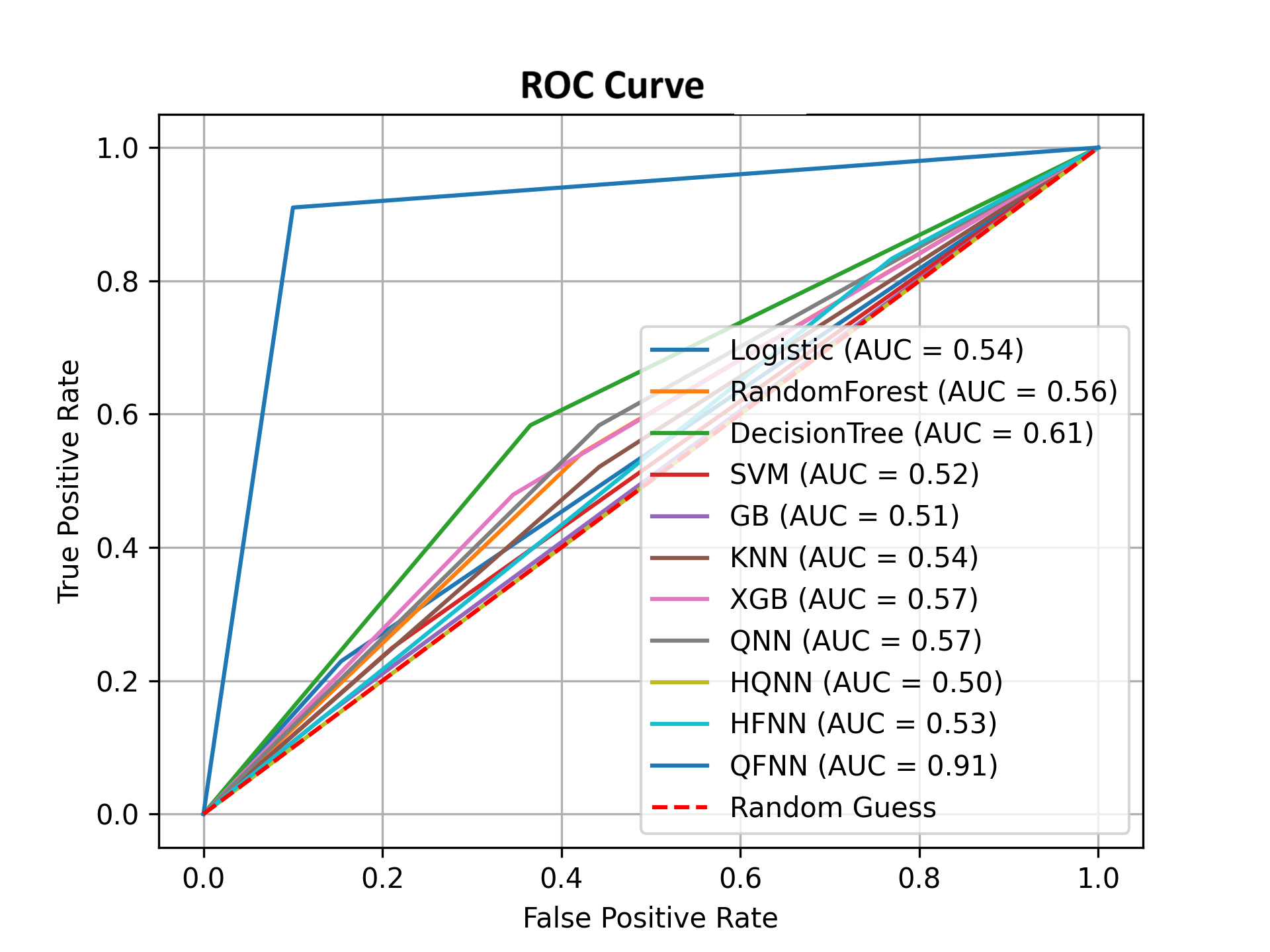}
\caption{}
\label{roc2}
\end{subfigure}\hfill
\caption{ROC curves with the AUC values for (a) CVTD and (b) GSTD.}
\label{ROC}
\end{figure}

\begin{figure}[!ht]
\centering
\begin{subfigure}{\linewidth}
\centering
\includegraphics[width=0.7\linewidth]
{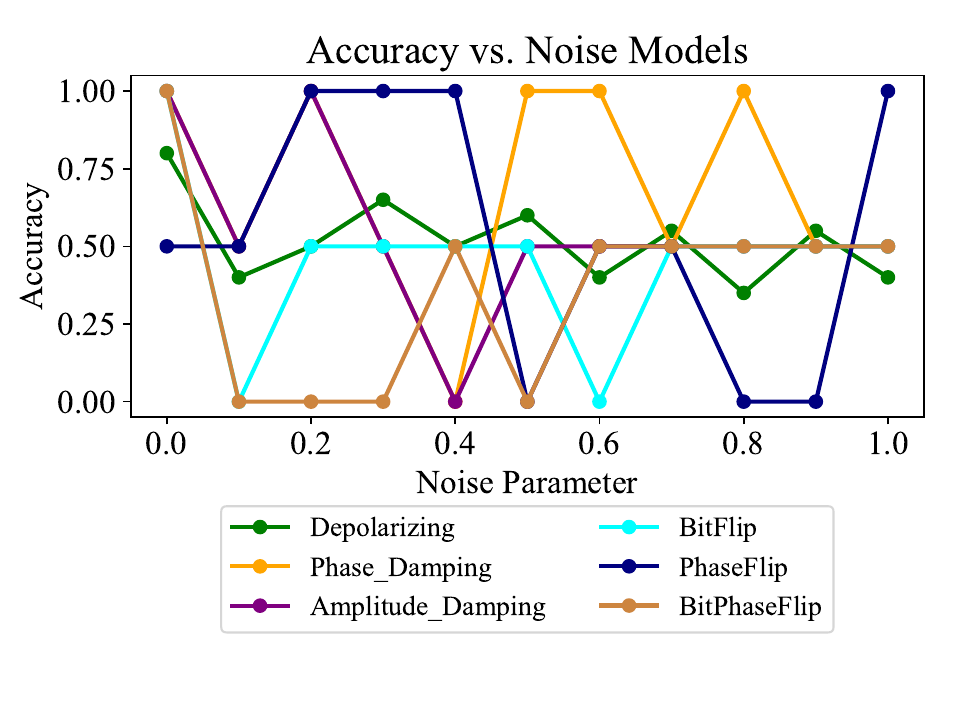}
\vspace{-0.6cm}
\caption{}
\label{Fig4f}
\end{subfigure}\hfill
\begin{subfigure}{\linewidth}
\centering
\includegraphics[width=0.7\linewidth]
{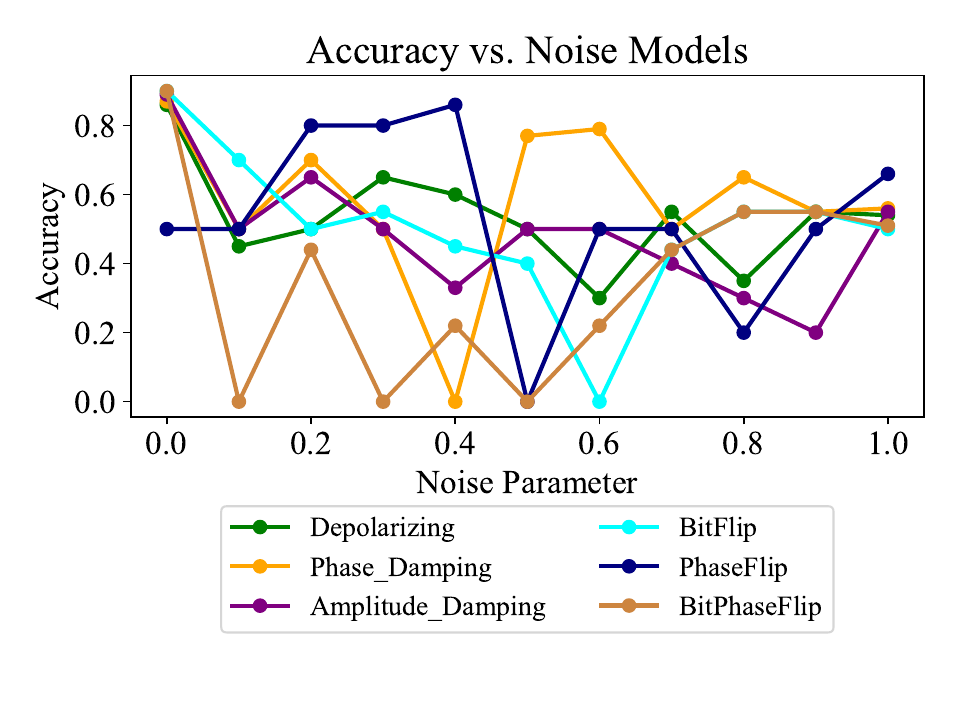}
\vspace{-0.6cm}
\caption{}
\label{Fig4f2}
\end{subfigure}\hfill
\caption{The accuracy of the QFNN in the presence of six noise models for the (a) CVTD and (b) GSTD.}
\end{figure}

\begin{table*}[ht]
\centering
\label{table:experimental_setup}
\begin{tabular}{|c|c|c|c|c|c|}
\hline
\textbf{Parameter} & \textbf{QFNN} & \textbf{QNN} & \textbf{ANN} & \textbf{HFNN}\\ 
\hline
\hline
{Qubits} & \textbf{2},4 & \textit{2},4 & N/A & 2,\textit{4} \\ 
\hline
{Optimizer} & \multicolumn{4}{c|}{\textit{Adam}} \\ 
\hline
{Learning Rate} & \multicolumn{4}{c|}{\textit{0.1}, 0.01, \textit{0.001}} \\ 
\hline
{Iterations/Epochs} & \multicolumn{2}{c|}{20,30,100} & 60 & {12, 15} \\ 
\hline
{Number of Parameters} & 16,\textit{19} & 8,\textit{11} & 7 (4 neurons + 1 output + 2 biases) & \textit{16},24 \\ 
\hline
{Simulator/Backend} & \multicolumn{2}{c|}{\textit{PennyLane}, IBM Quantum[Qiskit]} & N/A & \textit{Qiskit} \\ 
\hline
{Performance Metrics} & \multicolumn{4}{c|}{Accuracy, recall, precision, loss, F1 score, false negative (FN) rate, false discovery (FD) rate} \\ 
\hline
{Dataset Usage} & \multicolumn{4}{c|}{\textit{Two} datasets} \\ 
\hline
{Testing Split} & \multicolumn{4}{c|}{20\%, 30\%, \textit{50\%} of the dataset} \\ 
\hline
\end{tabular}
\caption{Detailed experimental setup and model parameters.}
\label{Table-2}
\end{table*}

\subsection{Datasets}
Two SA datasets are used to evaluate the performance of the proposed models; the first one is the CVTD \cite{Kaggledataset}, a collection of tweets about the COVID-19 pandemic. The sentiment classes include ``extremely negative'', ``negative'', ``neutral'', ``positive'', and ``extremely positive''. Notably, these five sentiment categories are transformed into a binary classification problem, categorizing tweets as positive or negative. The second dataset is a GSTD \cite{Kaggledataset2}, which comprises a collection of tweets used for SA. To understand the algorithm’s performance clearly, the multi-class sentiment targets are transformed into a binary classification format. 

\begin{table}[htbp]
    \centering
    \begin{tabular}{|c|c|c|}
\hline    \textbf{Model}  & \textbf{Loss (\%)} & \textbf{Accuracy (\%)} \\
   \hline 
   \hline 
    LR & -- & 79.78 \\
    \hline
    DT & -- & 75.59 \\
    \hline
    RF & -- & 75.80 \\
    \hline
    SVM & -- & 79.96 \\
    \hline
    NB & -- & 73.59 \\
    \hline
    KNN & -- & 78.61 \\
    \hline
    GB & -- & 79.75 \\
    \hline
    XGBoost & -- & 82\\
    \hline
    ANN & 0.42 & 81 \\
    \hline
    CF & -- & 51.25 \\
    \hline
    QSVM & -- & 60 \\
    \hline
    QNN & 0.42 &  50.3\\
    \hline
    QFNN & 0.11 & 100 \\
    \hline
    HQNN & 0.26 & 90.2 \\
    \hline 
    HFNN & 0.18 & 91.5 \\
    \hline 
    \end{tabular}
\caption{Comparison of accuracy and loss among classical, quantum, and hybrid methods for CVTD.}
\label{tab1}
\end{table}
\begin{table*}[htbp]
    \centering
    \begin{tabular}{|c|c|c|c|c|c|c|c|c|}
\hline    \textbf{Model} & \textbf{Accuracy (\%)}  & \textbf{Recall (\%)} & \textbf{Precision (\%)} & \textbf{F1 Score (\%)} & \textbf{FP Rate (\%)} & \textbf{FN Rate (\%)} & \textbf{FD Rate (\%)}\\
    \hline 
    \hline
     CF & 51.25  & 40.7 & 51.30 & 44 & 43.1  & 59.2 & 51.17 \\
    \hline     
         
    QFNN & 100 &  85 & 81 & 91 & 10.81 & 27.09 &  12.98  \\     
    \hline
    HFNN& 90.2  & 92.88  & 88.3 & 90.5  & 12.5 & 7.1  & 11.6  \\
    \hline
    \end{tabular}
\caption{Comparison of CF, QFNN and HFNN for CVTD against various accuracy metrics (accuracy, recall, precision, F1 score, FP rate, FN rate, FD rate.}
\label{tab2}
\end{table*}
\begin{table}[htbp]
    \centering
    \begin{tabular}{|c|c|c|}
\hline    \textbf{Model}  & \textbf{Loss } & \textbf{Accuracy (\%)} \\
   \hline 
   \hline 
    LR & -- & 79.60 \\
    \hline
    DT & -- & 69.80 \\
    \hline
    RF & -- &  74.40\\
    \hline
    SVM & -- & 77.80 \\
    \hline
    NB & -- & 58.60 \\
    \hline
    KNN & -- & 67.40 \\
    \hline
    GB & -- &  75\\
    \hline
    XGBoost & -- & 73\\
    \hline
    ANN & 0.64 &  77\\
    \hline
    CF & -- &  50.78\\
    \hline
    QSVM & -- & 42.33 \\
    \hline
    QNN &  0.41 &  52.80  \\
    \hline
    QFNN & 0.20 & 90 \\
    \hline 
    HQNN & 0.69  & 50 \\
    \hline
    HFNN &  0.25 & 47 \\
    \hline
    \end{tabular}
\caption{Comparison of accuracy and loss among classical, quantum, and hybrid methods for GSTD.}
\label{tab2nddataset}
\end{table}

\begin{table*}[htbp]
    \centering
    \begin{tabular}{|c|c|c|c|c|c|c|c|c|}
\hline    \textbf{Model} & \textbf{Accuracy (\%)}  & \textbf{Recall (\%)} & \textbf{Precision (\%)} & \textbf{F1 Score (\%)} & \textbf{FP Rate (\%)} & \textbf{FN Rate (\%)} & \textbf{FD Rate (\%)}\\
    \hline 
    \hline
    CF & 50.78  & 69.8 & 51.5 & 57.9 & 71.4 & 30.1  & 50.4  \\
    \hline          
    QFNN & 90 & 92  &  90 &  85 & 07.64 & 37.33 &  29.64  \\
    \hline 
    HFNN & 47 & 73.8  & 53.7 & 62.1  & 69.9 & 26.1  & 46.2  \\
    \hline
    \end{tabular}
\caption{Comparison of CF, QFNN, and HFNN for GSTD against various accuracy metrics (accuracy, recall, precision, F1 score, FP rate, FN rate, FD rate.}
\label{tab3}
\end{table*}
\begin{table}[h]
    \centering
\begin{tabular}{|c|c|c|}
\hline
\textbf{Model} & \textbf{Datasets} &\textbf{Accuracy (\%)} \\
\hline
\hline
QSVC and VQC \cite{bib_Masum} & Bengali& 72.2 \\
\hline
QSVC and VQC \cite{bib_Masum} & Twitter & 71.2\\
\hline
QSVM \cite{bib_Omar}& Arabic tweets & 92.0 \\
\hline
CFN \cite{bib_Zhang_CFN} &  MUStARD& 75.4 \\
\hline
CFN \cite{bib_Zhang_CFN} &  Reddit track& 68.0 \\
\hline
\textbf{QFNN }  &  \textbf{CVTD} &  \textbf{100} \\
\hline
\textbf{QFNN}  &  \textbf{GSTD} & \textbf{90} \\
\hline
\textbf{HFNN}  &  \textbf{CVTD} &  \textbf{90.2} \\
\hline

\end{tabular}
\caption{Comparison between existing quantum models and our approaches for classification of various SA datasets.}
    \label{qml}
\end{table}

\subsection{Preprocessing}
The SA methodology begins with systematically starts by systematically preprocessing the textual data for analysis, as shown in Fig. \ref{figpreprocess}. This involves standard procedures such as tokenization, lowercasing, removing stop words, stemming/lemmatization, and other normalization techniques. Once the text is prepared, the relevant features (i.e., significant terms or phrases that indicate sentiment) are identified, and irrelevant or redundant features are removed to improve efficiency following the steps shown in Fig. \ref{figtweets}. The problem is framed as a binary classification task, where sentiments are categorized into two distinct classes, A and B. Neutral sentiments are excluded, focusing only on clear positive or negative sentiments. Classes A and B are numerically encoded, typically with labels 1 for class A and 0 for class B. NLP techniques are employed to enhance the quality and relevance of the textual data, and a word frequency dictionary is constructed for each sentiment class (A and B), where the frequency of words in each class is counted. This creates a numerical representation that our algorithms can use to classify the text into sentiment A or B.

\subsection{Metrics and Hyperparameters}
The study evaluates classical, quantum, and hybrid fuzzy algorithms on two datasets using various performance metrics. Quantum circuits with 2 and 4 qubits are used to build QNN, HQNN, and HFNN, with a reduced feature map constructed using the “ZZFeatureMap” with three repetitions and a “full” entanglement connection. EfficientSU2 ansatz with 16 parameters is found to perform better than RealAmplitude in terms of accuracy. Fine-tuning HQNN and HFNN involves testing different neural network structures and configurations to optimize performance. The experiments, detailed in table \ref{Table-2}, include testing various qubits, learning rates, and loss functions (MSE and cross-entropy), with the training dataset containing 1000 data points and 500 for testing.

\subsection{Noise Models}
{Noise models are theoretical frameworks that explain the impact of noise and errors on quantum systems. One powerful tool for describing these models is Kraus operators, which is a set of operators capable of representing any form of quantum error channel. This mechanism takes a quantum state as input and outputs another with a specific error probability. We verified the robustness of the proposed QFNN algorithm in the presence of six noise models: bit flip (BF), phase flip (PF), bit phase flip (BPF), depolarizing (DP), amplitude damping (AD), and phase damping (PD) \cite{bib_Kumar}.}

\subsection{Results and Analysis}
The performance analysis of the proposed models for CVTD and GSTD are tested and compared against various classical ML models such as LR, DT, RF, SVM, NB, KNN, GB, XGBoost, ANN, and CF, quantum models such as QNN and QSVM (tables \ref{tab1} and \ref{tab2nddataset}). A comparison among the existing quantum models used for the classification of SA datasets is presented in table \ref{qml}. The dataset used for training includes 1000 data points, and 500 data points are reserved for testing. The quantum and hybrid algorithms are run using qiskit and the ``qasm'' quantum simulator. 
\subsubsection{CVTD}
The classical ML models achieve accuracies between 73.59\% and 79.96\%, with the ANN reaching 81\% ((Fig. \ref{Fig4a}). In contrast, the QNN shows the lowest accuracy at 50.3\% \ref{Fig4b}), while the HQNN significantly improves to 90.2\%, the accuracy starts at around 50\% and increases exponentially with an increase in epoch number (Fig. \ref{Fig4c}). In terms of loss, HQNN has the lowest loss at 26.4\%, followed by ANN at 42.7\%, and QNN at 37.4\%. This demonstrates that the addition of a hybrid component to QNN enhances both accuracy and reduces loss. Notably, the HFNN achieves 91.5\% accuracy with a minimal loss of 18.8\% (Fig. \ref{Fig4e}), and the QFNN achieves an outstanding 100\% accuracy with a corresponding loss of 11\%, the lowest among all models (Fig. \ref{Fig4d}). These results, summarized in table \ref{tab2}, highlight the superior performance of hybrid and quantum-fuzzy models, particularly on the CVTD dataset.

\subsubsection{GSTD}
The classical ML models achieve accuracies ranging from 50.78\% to 79.60\%, with LR performing the best among them at 79.60\%. The QSVM method shows the lowest accuracy at 42.33\%, while the QFNN achieves the highest accuracy at 90\%. In terms of loss, QFNN also records the lowest loss at 20.96\%, followed by the HFNN with a loss of 25\%, and the HQNN with a higher loss of 69.4\%. This indicates that QFNN leads both in accuracy and loss reduction. For ANN, the accuracy stabilizes at around 71\% (Fig. \ref{Fig4a2}), while QNN's accuracy remains constant at 50\%, although its loss decreases to below 45\% after five epochs (Fig. \ref{Fig4b2}). HQNN shows fluctuating accuracy around 50\% (Fig. \ref{Fig4c2}), while HFNN achieves 47\% accuracy with a 25\% loss (Fig. \ref{Fig4e2}). Remarkably, QFNN reaches an impressive 90\% accuracy with a loss of 20.96\%, outperforming all other algorithms (Fig. \ref{Fig4d2}). The performance metrics for these algorithms on the GSTD are summarized in table \ref{tab3}. 

\subsection{Effect of Noise}
The robustness and accuracy of the QFNN under different noise models for the CVTD and GSTD are verified in Figs. \ref{Fig4f} and \ref{Fig4f2}, respectively. For CVTD, the DP error model demonstrates the highest robustness, maintaining non-zero accuracy even at the highest noise levels, while other models' accuracy drops to 0\% between noise parameters of 0.1 and 0.9. The accuracy of these other models fluctuates unpredictably with noise, making mitigation more difficult compared to the DP error model. For GSTD, error models exhibit more complex behavior, with medium accuracy persisting even at maximum noise levels, indicating that these models can learn to mitigate extreme noise. Furthermore, the BPF model is highly sensitive to specific noise types, showing high accuracy at a noise parameter of 0.9, dropping to 0\% at 0.4, and rising to medium accuracy at maximum noise. These patterns suggest that some error models learn to exploit noise structure, particularly when noise exhibits correlation, allowing them to mitigate its effects. The DP and AP error models never drop to 0\% accuracy, unlike the other models, which all hit 0\% at noise parameters between 0.4 and 0.6.


\section{Discussion and Conclusion\label{SecV}}
 
In this paper, we introduce a novel hybrid approach for SA that combines quantum computing and neuro-fuzzy systems. This hybrid model is designed to address the challenges faced by classical neuro-fuzzy algorithms. By leveraging principles from quantum mechanics, our approach formulates a model that surpasses the capabilities of individual classical and quantum models. Additionally, the incorporation of fuzziness adds greater flexibility to explore the space of combinations and adapt the model boundary more effectively. To evaluate the effectiveness of our proposed approach, sentiments are analyzed, tested, and evaluated using various classical algorithms (LR, DT, RF, SVM, NB, KNN, GB, XGBoost, ANN, and CF), quantum algorithms (QNN, QSVM, and QFNN), and hybrid algorithms (HQNN and HFNN) on two Twitter datasets: CVTD and GSTD. The findings showed that QFNN outperformed all classical, quantum, and hybrid algorithms, achieving an accuracy of 100\%, and the HQNN method attained an accuracy of 90.2\%. Among the classical techniques, ANN achieved the highest accuracy at 81\%, while the quantum methods, QSVM and QNN, yielded 60\% and 50.3\% accuracy, respectively. The hybrid model demonstrated a 10\% improvement over nine classical methods and a 48\% improvement over the QNN method. QFNN's performance excelled in CVTD (100\%) and GSTD (90\%).  This is further supported by the ROC curves and corresponding AUC values, as shown in Fig. \ref{ROC}. Despite CF ranking second lowest in CVTD and third lowest in GSTD, the infusion of quantum nature elevated its accuracy to the highest. Consistency was observed in QFNN's delivery of higher accuracy across both datasets, even when subjected to six noisy channels (BF, PF, BPF, DP, AD, and PD).

A comparison of various quantum models for SA across different datasets revealed that the QFNN model exhibited remarkable performance, achieving a perfect accuracy of 100\%. Notably, QFNN demonstrated exceptional stability against various noise models, with the DP channel identified as the optimal choice. This ensured that QFNN maintained non-zero accuracy levels, highlighting its robustness in practical applications. The integration of a fuzzy layer further enhanced the accuracy of QFNN, surpassing hybrid and CF approaches. The enhanced flexibility and adaptability provided by the fuzzy layer enabled QFNN to navigate complex data and uncertain boundaries effectively and with greater precision. Consequently, QFNN demonstrated its capability to accurately process and analyze a growing volume of textual data, effectively handling noise, outliers, and the complexities of language structures, thereby improving sentiment classification. This positions QFNN as a potentially promising solution in the field of SA, with implications extending beyond natural sciences to social sciences. This research is presently restricted by the utilization of a binary dataset, thereby constraining the application of models to multi-class datasets. Implementing enhanced error mitigation techniques in the presence of quantum hardware is imperative. Further research is necessary to encompass a range of error mitigation techniques to facilitate hardware implementation in sentiment analysis applications. Additionally, focused research efforts are required to enhance computational efficiency through GPU utilization, particularly when managing large datasets and increasing qubit sizes within this context.

\section*{Acknowledgment}
Work by N.I was supported in part by the NYUAD Center for Quantum and Topological Systems (CQTS), funded by Tamkeen under the NYUAD Research Institute grant CG008.

\bibliographystyle{IEEEtran}

\bibliography{IEEE}

\end{document}